\documentclass[10pt,twocolumn,letterpaper]{article}

\usepackage[pagenumbers]{cvpr} 

\usepackage{bbding}
\usepackage{pifont}
\usepackage{wasysym}
\usepackage{amssymb}
\usepackage{xcolor} 
\usepackage{microtype}
\usepackage{graphicx}
\usepackage{booktabs} 
\usepackage{multirow}
\usepackage{amsmath,amssymb}
\usepackage{booktabs}
\usepackage{tabularx}
\usepackage{tcolorbox}
\usepackage{array}
\usepackage{caption,subcaption}
\newcommand{\Xmat}[0]{{{\bf X}}}
\newcommand{\thetav}{\boldsymbol{\theta}}
\newcommand{\xv}{\boldsymbol{x}}
\definecolor{mygreen}{HTML}{006400}
\definecolor{cvprblue}{rgb}{0.21,0.49,0.74}
\usepackage[pagebackref,breaklinks,colorlinks,allcolors=cvprblue]{hyperref}
\makeatletter
\def\blfootnote{\gdef\@thefnmark{}\@footnotetext}
\makeatother
\newcommand{\sks}{\texttt{<sks>}}

\def\paperID{2878} 
\def\confName{CVPR}
\def\confYear{2025}

\title{Yo'Chameleon:\\Personalized Vision and Language Generation}

\author{
Thao Nguyen$^{1,2}$ \quad Krishna Kumar Singh$^{2}$ \quad Jing Shi$^{2}$ \quad Trung Bui$^{2}$ \quad Yong Jae Lee$^{1,\P}$ \quad Yuheng Li$^{2,\P}$\\
$^{1}$University of Wisconsin--Madison \quad $^{2}$Adobe Research \\
{\small \url{https://thaoshibe.github.io/YoChameleon}
}
}

\makeatletter
\let\@oldmaketitle\@maketitle

\renewcommand{\@maketitle}{\@oldmaketitle
\vspace{-6mm}
\centering
\includegraphics[width=1\linewidth]{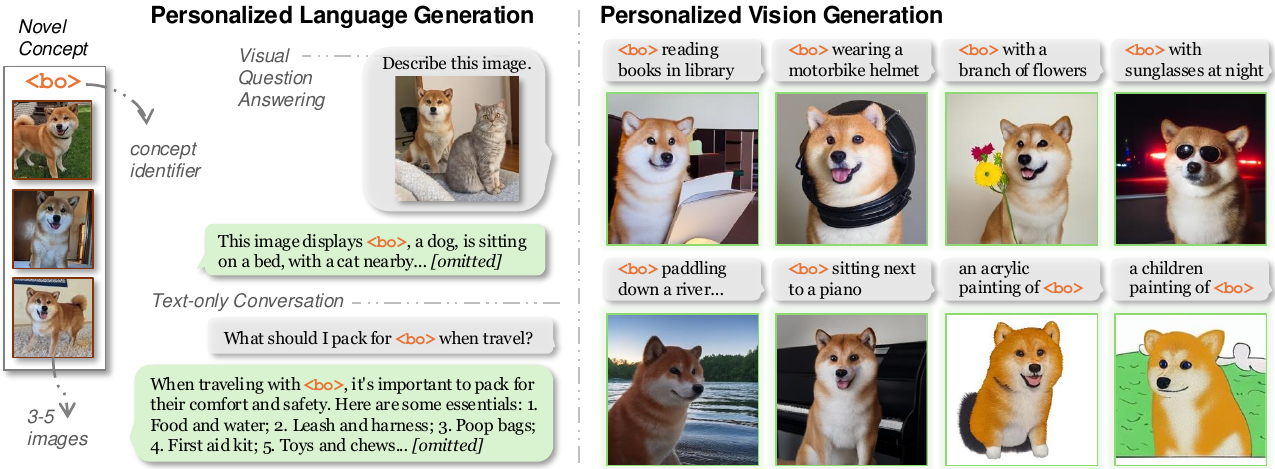}
\vspace{-5mm}
\captionof{figure}{Using only 3-5 images of a novel concept/subject, we personalize Large Multimodal Models (e.g., Chameleon~\cite{chameleon}) so that they retain their original capabilities while enabling tailored language and vision generation for the novel concept.
}
\label{fig:feature-graphic}
\vspace{3mm}
}
\makeatletter

\begin{document}
%
%
%
%

\maketitle
\blfootnote{$\P$ denotes equal advising}
\vspace{-6mm}
\begin{abstract}
\vspace{-6mm}

Large Multimodal Models (e.g., GPT-4, Gemini, Chameleon) have evolved into powerful tools with millions of users.
However, they remain generic models and lack personalized knowledge of specific user concepts.
Previous work has explored personalization for text generation, yet it remains unclear how these methods can be adapted to new modalities, such as image generation.
In this paper, we introduce Yo'Chameleon, the first attempt to study personalization for large multimodal models.
Given 3-5 images of a particular concept, Yo'Chameleon leverages soft-prompt tuning to embed subject-specific information to (i) answer questions about the subject and (ii) recreate pixel-level details to produce images of the subject in new contexts. Yo'Chameleon is trained with (i) a self-prompting optimization mechanism to balance performance across multiple modalities, and (ii) a ``soft-positive" image generation approach to enhance image quality in a few-shot setting.
Our qualitative and quantitative analyses reveal that Yo'Chameleon can learn concepts more efficiently using fewer tokens and effectively encode visual attributes, outperforming prompting baselines.
\end{abstract}

\vspace{-5mm}
\section{Introduction}
\label{sec:intro}


Recent advances in Large Multimodal Models (LMMs) have transformed them into versatile, general-purpose AI assistants~\cite{gpt4o,chameleon,emu3,gemini,transfusion,wu2024janus}. These models are increasingly being integrated into everyday applications, offering enhanced performance, improved efficiency, and support for multiple modes of communication. The ability to process both visual and textual information within a single system—as demonstrated by models like GPT-4o~\cite{gpt4o} and Gemini~\cite{gemini}—has streamlined user interactions and improved query comprehension.

Modern LMMs enable seamless two-way communication through both text and images. Users can input queries combining natural language and visual elements, and the models can respond with both textual descriptions and generated images. For instance, when asked to ``Describe a Shiba Inu dog and generate a photo of it'', these AI assistants can now provide comprehensive responses that combine detailed descriptions with visual representations.

While LMMs excel at general tasks, they face limitations when handling personalized queries. For example, if asked ``Can you describe \texttt{<bo>} and generate a photo of \texttt{<bo>} reading books in library?" these models cannot provide accurate responses without prior knowledge of the specific pet (e.g., a dog named \texttt{<bo>}).
This highlights a crucial gap in their capabilities, as human interaction with the world is inherently personal --- we engage with our own devices, pets, friends, and environments. To create more meaningful AI interactions, LMMs need mechanisms to learn, understand, and generate user-specific concepts, enabling them to evolve from general-purpose tools into personalized assistants (Fig.~\ref{fig:feature-graphic}).

Personalization techniques have been extensively studied for LLMs~\cite{lamp,zhang-etal-2018-personalizing,liu-etal-2023-recap,gjurkovic-etal-2021-pandora,jang2023personalizedsoupspersonalizedlarge} and image generation models~\cite{ruiz2023dreambooth,textual_inversion,instantbooth,he2024imagineyourselftuningfreepersonalized,ye2023ip-adapter,subjectdiffusion,kumari2022customdiffusion}, demonstrating significant progress in these individual domains. Recent works~\cite{yollava, myvlm,captioningremember} have begun exploring personalization for vision-language models like LLaVA, which can take both image and text as inputs but only generate textual outputs. Despite this progress, the challenge of personalizing LMMs --- which require both personalized text/image understanding and generation capabilities remains largely unexplored. In this paper, we identify two key challenges in extending personalization to these more comprehensive multimodal systems. To be specific, we focus on Large Multimodal Models that capable of understanding and generating images and text (e.g., Chameleon~\cite{chameleon}).

The first challenge is catastrophic forgetting. Image generation tasks require granular information of new concepts, typically necessitating part/full model fine-tuning to achieve satisfactory results (e.g.,~\cite{ruiz2023dreambooth}). However, LMMs store both visual and textual information, and our empirical studies show that fine-tuning for image generation (e.g., similar to~\cite{ruiz2023dreambooth}) causes the model to rapidly lose its world knowledge, compromising its functionality as a general-purpose AI assistant. Conversely, soft prompt learning~\cite{prompt_tuning,hao2023toolkengpt}, which introduce learnable tokens to encode new concepts while keeping the model frozen, is effective for personalized image understanding tasks~\cite{yollava}. Although, our experiments reveal that soft prompt learning with only 3-5 user images fails to produce high-quality image generation results.

To address this challenge, we first identify that the limited number of training images is a key factor preventing soft prompt learning from matching full model fine-tuning's performance. Our study demonstrates that with $\sim$300 real images of a concept, soft prompt learning can achieve comparable performance to full-model fine-tuning while preserving the LMM's pretrained knowledge. However, since users typically only provide 3-5 images for a new concept (positive images), this is not a practical solution. Drawing from this analysis, we propose leveraging ``soft-positive'' images that share significant visual similarities with positive samples to enrich the training data. To effectively utilize these ``soft positive'', we implement an adaptive prompt length strategy where the prompt length varies based on the visual similarity between ``soft-positive'' and positive samples. For instance, when training the model to recognize a user's Shiba Inu, we utilize images of similar-looking Shiba Inu with adaptive prompt lengths to augment the limited training data. The more similar the ``soft-positive'' image is to the real positive images, we will use longer soft prompt to describe it.

The second challenge is the incompatibility between image generation and understanding capabilities within LMMs. Our experiments reveal that soft prompt optimized for one task cannot effectively transfer to the other. Specifically, when soft prompt trained for image understanding (text generation) are used to for image generation, the LMM produces irrelevant visual content. This phenomenon aligns with prior work~\cite{hao2023toolkengpt,visii,hard_prompt} suggesting that optimized textual representations for one task might not be interpretable. Jointly training the soft prompt on both tasks might seem like an intuitive solution, however, our empirical results show this approach leads to suboptimal performance for both tasks (Fig.~\ref{fig:multi-tasks}).

To enable effective personalization across both tasks, we propose using dual soft prompts --- one specialized for text generation and another for image generation. This approach demonstrates superior performance compared to using a single set of prompts. Additionally, we introduce a self-prompting mechanism where the model first determines the task type (i.e., understanding or generation) before responding to queries, allowing it to better utilize the appropriate set of prefix tokens for each task.

In summary, our contributions are: 

\begin{itemize}
    \item We introduce the first attempt of personalization with Large Multimodal Models (i.e., models that capable of understanding and generating images and text).
    \item We present a novel ``soft-positive'' concept with dynamic prompt length to enhance the image generation quality.
    \item We propose a novel approach, in which use two set of soft prompts and self-prompting optimization techniques to balance the performance across the modality.

\end{itemize}


\section{Related Work}
\label{sec:related work}


\paragraph{Personalization for Large Multimodal Models.} 
Large Language Models (LLMs)~\cite{llama,llama2,jiang2023mistral7b,brown2020languagemodelsfewshotlearners} and text-to-image models~\cite{rombach2021highresolution,dalle,shi2020improvingimagecaptioningbetter,saharia2022photorealistictexttoimagediffusionmodels,li2023gligen} have made tremendous progress recently, demonstrating extensive knowledge and excelling at text and image generation, respectively. Vision-language models~\cite{liu2024llavanext,liu2023improvedllava,liu2023llava,chen2023internvl,chen2024far,gao2024mini} have emerged as a bridge between these modalities, capable of processing image-text inputs and generating textual outputs. Building upon this, researchers have developed unified Large Multimodal Models (LMMs)~\cite{transfusion,gemini,gpt4o,emu3,chameleon} that capable of understanding and generating both images and text.

However, these foundational models typically possess generic knowledge, making personalization a crucial and active research area. For LLMs, personalization often involves storing descriptions of personalized subjects as prompts in databases for reference during user interactions~\cite{liu-etal-2023-recap,retrieval_chatbot,lamp}. In image generation, researchers typically fine-tune either the entire model or specific components to incorporate visual knowledge~\cite{ruiz2023dreambooth,ye2023ip-adapter,lin2024ctrlx,instantbooth,he2024imagineyourselftuningfreepersonalized,wei2023elite,visii}. Recent work by~\cite{yollava,myvlm} proposes personalizing vision-language models through soft prompts to enable recognition and discussion of user-specific objects. Despite these advances, personalization of unified image/text generation models remains unexplored. Our work addresses this gap by investigating the challenges and potential solutions in this emerging area.
\vspace{-3mm}
\paragraph{Parameter-Efficient Fine-Tuning (PEFT).} Fine-tuning large pretrained models is often suboptimal due to computational costs and the risk of catastrophic forgetting. Consequently, numerous PEFT methods have been introduced to optimize a small subset of parameters (or introduce extra parameters) for downstream tasks~\cite{lora,li-liang-2021-prefix,prompt_tuning}. In the domain of LLMs, prompt tuning (or soft-prompts) has emerged as an effective approach to adapt pretrained language models for various tasks, such as tool utilization~\cite{hao2023toolkengpt} and text classification~\cite{prompt_tuning}. This approach has recently been extended to personalize vision-language models~\cite{yollava}.
However, existing vision-language model approaches (e.g.,~\cite{yollava}) primarily focus on text generation objectives. Our experiments reveal that naively extending their soft-prompting approach to encompass both text and image generation yields suboptimal results, as these tasks are not naturally complementary. To this end, we propose a self-prompting technique where the model first predicts the task type before generating the response. This approach effectively resolves the challenges of personalizing models with multi-modal outputs.
\vspace{-3mm}
\paragraph{Hard negative image mining.} Negative images have been widely used in the computer vision community~\cite{object_detection,xuan2021hardnegativeexampleshard,wan2016bootstrappingfacedetectionhard,deng2018arcface}. In vision-language model personalization, \cite{yollava} employs this technique to enhance personal object recognition. For image generation personalization, \cite{ruiz2023dreambooth,kumari2022customdiffusion} utilize negative examples as regularization to prevent model forgetting of class-level information. SuTI~\cite{suti} and COTI~\cite{coti} leverage negative images that are visually similar to personalized objects to establish a better initialization that facilitates easier adaptation to the target personalized object. However, unlike them which treats all negative images equally, we pursue a more nuanced approach. We propose an adaptive soft prompt length mechanism based on the visual similarity between negative images and positive examples. Specifically, we treat these negative images as ``soft-positive'' examples, allocate more prompt length to ``soft-positive'' images that exhibit higher visual similarity to the positive examples, allowing for more fine-grained representation learning.

\begin{figure}[t]
    \centering
    \includegraphics[width=1\linewidth]{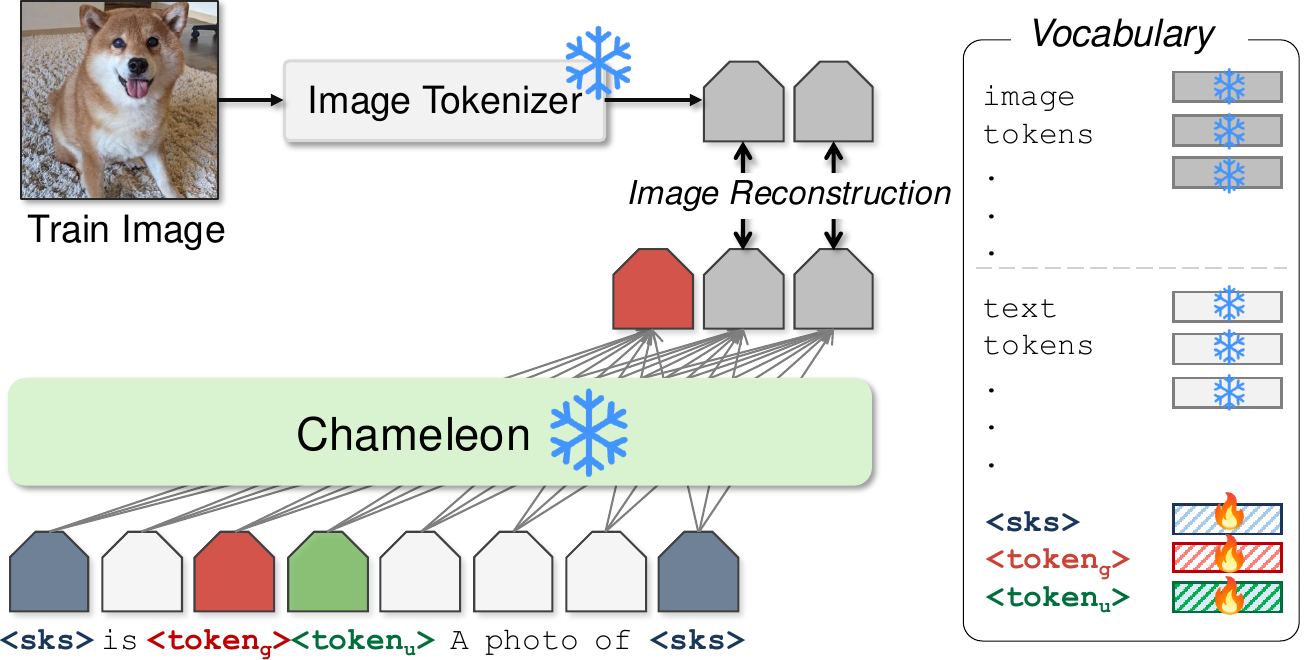}
    \caption{Image Reconstruction. The generated image, conditioned on a personalized prompt, is compared with the ground truth image to calculate the image reconstruction loss.}
    \label{fig:img-reconstruction}
    \vspace{-5mm}
\end{figure}

\section{Yo'Chameleon}
\label{sec:approach}

Given a handful of images of a concept that we want to learn $I^{1}, I^{2}, ..., I^{n}$ (typically 3-5 images), our goal is to enable LMMs (i.e., Chameleon~\cite{chameleon}) to embed the concept into a special token (e.g., \sks) and to perform: (1) Personalized language generation (e.g., ``Describe \sks''; or given an image, ``Where is \sks~in this image?''); and (2) Personalized vision generation (e.g., ``Generate a photo of \sks'').

We first present how to present novel concept for as learnable prompt for LMMs (Sec.~\ref{sec:representing}). Subsequently, we outline how to achieve personalized image generation (Sec.~\ref{sec:image_generation}). Finally, we discuss how to unify both image generation and understanding capabilities within a single model (Sec.~\ref{sec:self_prompt}). As we chose Chameleon~\cite{chameleon} as our base model, we named our method Yo'Chameleon, with \textit{Yo} (short for \textit{Your}) adopted from Yo'LLaVA's~\cite{yollava} personalization of LLaVA~\cite{liu2023llava}.

\subsection{Representing a Concept as a Learnable Prompt}
\label{sec:representing}
In image generation, prior work demonstrated that prompt tuning can effectively encode visual concepts for personalization~\cite{textual_inversion,hard_prompt,visii}. This success has extended to vision-language models, where studies like~\cite{hao2023toolkengpt,yollava} show that prompt tuning can effectively encode novel visual attributes for text-only generation and image understanding. Building on this paradigm, we propose to represent personalized subjects as learnable prompts for LMMs:
\begin{center}
    ``\texttt{\textcolor{blue}{<sks>} is \textcolor{red}{<$\text{token}_{1}$>}\textcolor{red}{<$\text{token}_{2}$>}$\dots$\textcolor{red}{<$\text{token}_{k}$>}}.''
\end{center}
where \sks~is a learnable unique identifier for this new concept, and \texttt{<$\text{token}_{i}$>} are the learnable tokens which should encode visual information of that concept.
This approach offers computational efficiency by only requiring updates to a small
subset of parameters (i.e., tokens) while preserving the original core model weights.

In the context of Chameleon~\cite{chameleon}, a model that we choose to build upon in this work, an image is broken down into a series of image tokens, wrapped by special tokens which indicate the start-of-image \texttt{<soi>} and the end-of-image \texttt{<eoi>}.
The training objective for both image and text remains consistent with standard autoregressive modeling, where the model learns to predict the next token in the sequence conditioned on the previous tokens.
Thus, training for personalization follows an instruction-tuning paradigm, where the loss computation is specifically focused on the response portion of the instruction-response pairs.
Given the conversation pair $(\Xmat_{\texttt{q}}^i, \Xmat_{\texttt{a}}^i)$, where $\Xmat_{\texttt{q}}^i$ is the question, and $\Xmat_{\texttt{a}}^i$ is the corresponding answer, the masked language modeling loss for each conversation of length $L$ by:
\begin{equation}
    p( \Xmat_{\texttt{a}}) =
    \prod_{j=1}^{L} p_{\thetav} (  \xv_j
| \Xmat_{\texttt{a}, <j}),
    \label{eq:auto_regressive}
\end{equation}
where $\Xmat_{\texttt{a}, <j}$ are the instruction and answer tokens in all turns before the current prediction token $\xv_j$. $\thetav$ is the trainable parameters, in this case, including the concept identifier \sks, $k$ latent tokens, and the final classifier head matrix $W$ of the language model that associated with these tokens.

\begin{figure}
    \centering
    \includegraphics[width=0.9\linewidth]{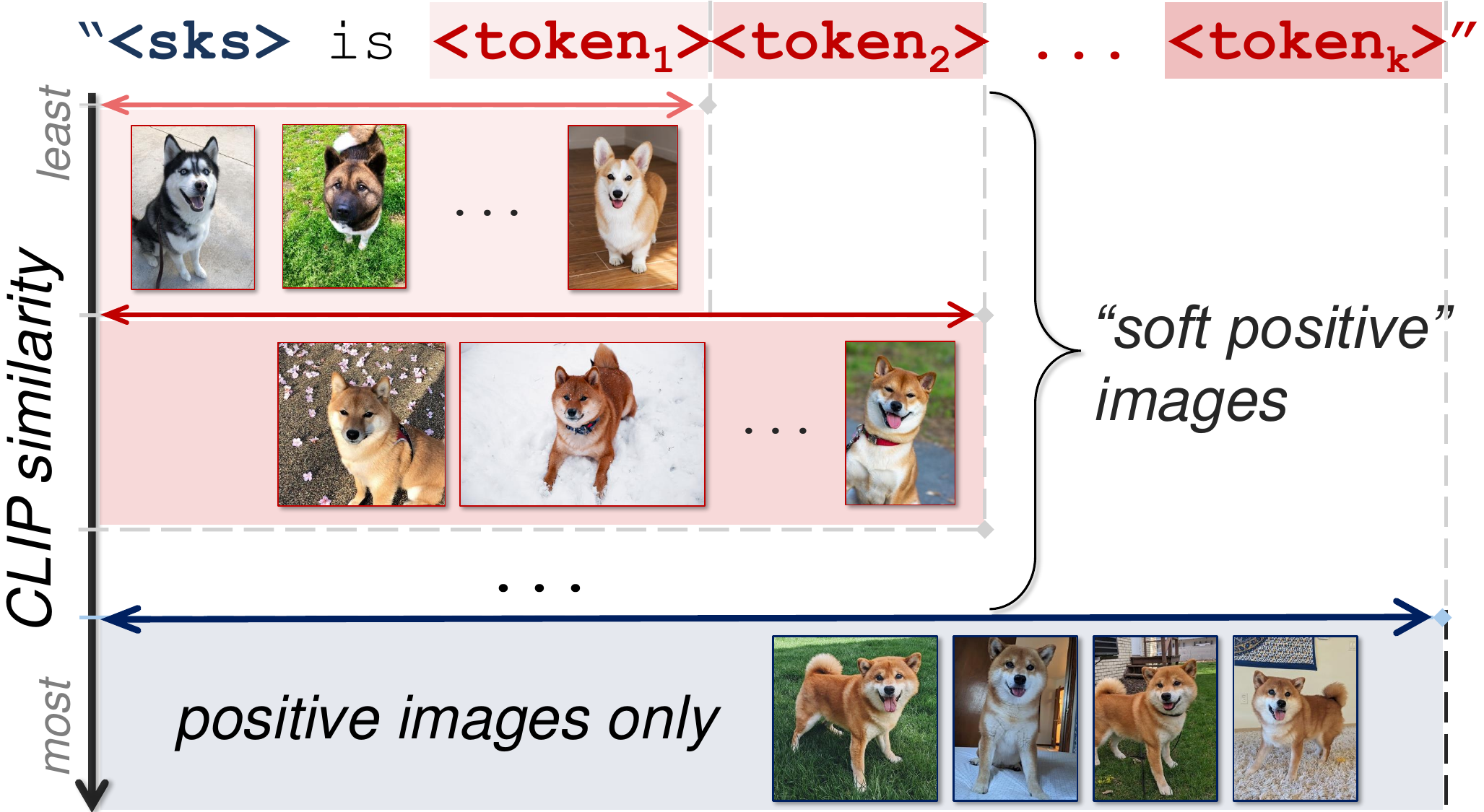}
    \caption{``Soft positive'' images. Retrieved images are ranked according to their similarity to positive images using CLIP image similarity scores. Images that are more similar to the actual positive images are described with more latent tokens (i.e., more details).}
    \label{fig:soft-positive}
    \vspace{-3mm}
\end{figure}

\subsection{Personalizing Image Generation}
\label{sec:image_generation}
A straightforward approach to personalization would be to directly train soft prompt on a limited set of $n$ positive images. However, optimization with such limited data often yields suboptimal results. To this end, researchers have explored two primary approaches to expand the training samples: (1) data augmentation  (e.g., background inpainting) to treat augmented images as additional positive training samples~\cite{anythinganywhere,svdiff,subjectdiffusion}, and (2) leveraging hard-negative samples as an initialization, in which we first train on these negative images, then add an additional step to fine-tune the results with a limited number of positive examples to enhance personalization~\cite{celebbasis,suti,coti}. Empirical evidence suggests that utilizing real negative examples produces superior results compared to synthetic data augmentation approaches.

Motivated by this, in our approach, we retrieve hard-negative images, but unlike prior work~\cite{suti,coti}, we use them as ``soft positive'' images. The key insight is that hard negative images can share varying degrees of similar characteristics with the positive samples, and thus should contribute differently to the learning process.
Taking the same example of an user's pet (a Shiba Inu) again, in this scenario, each negative image can function as a ``soft positive'' to varying extents. The similarity ranges from less to more similar negative images: for example, ``A dog'' (least similar), followed by ``A dog with a yellow coat'' (more similar), and so on.

Specifically, given $N$ retrieved negative images, we rank them from most to least similar to the average feature of the positive samples (based on CLIP image similarity score~\cite{clip}); Then, we divide them into $k-1$ groups, each containing roughly $N/(k-1)$ images, according to their ranking. During training, we implement an adaptive token allocation strategy: negative images with higher similarity scores are assigned more learnable tokens, allowing for more detailed representation of relevant features. The complete set of tokens is reserved exclusively for the true positive images \sks, ensuring that the model maintains the ability to distinguish the target concept while leveraging relevant features from similar soft positive examples. Fig.~\ref{fig:soft-positive} illustrates this hierarchical token allocation strategy.




\begin{figure}
    \centering
    \includegraphics[width=0.99\linewidth]{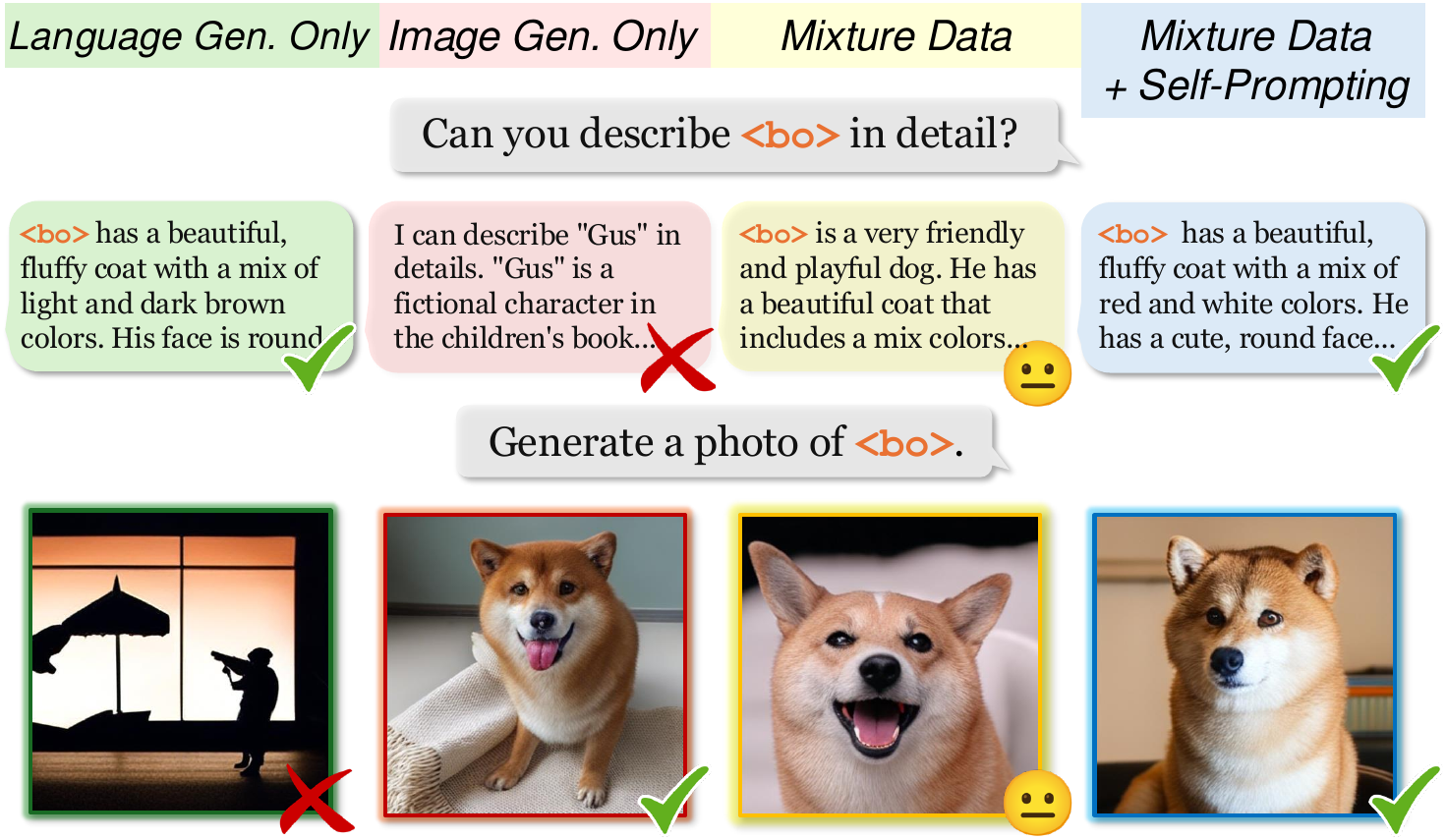}
    \caption{Optimized tokens for one task cannot effectively perform another, and simply training on a mixture of data yields suboptimal performance across tasks. We propose a self-prompting approach, where the model predicts which task to perform first, achieving the best of both worlds. (Input images are given in Fig.~\ref{fig:feature-graphic}).}
    \label{fig:multi-tasks}
    \vspace{-4mm}
\end{figure}

\subsection{Personalizing Image and Language Generation}
\label{sec:self_prompt}
For text generation and image understanding tasks, we follow the approach in~\cite{yollava} to create a training dataset, which comprises two primary components: recognition data and question-answering data. For recognition data, we construct a balanced dataset containing the handful of positive examples alongside both 100 easy negative examples and 100 hard negative examples. For question-answering, we adopt the template in~\cite{yollava}, which includes 10 questions (e.g., ``What type of object is \texttt{<sks>}?''). We use GPT-4o~\cite{gpt4o} to generate answers for these questions using the positive images.

To achieve our goal of personalizing LMMs for both text and image generation capabilities, a straightforward approach would be to simultaneously train soft prompt using both the understanding data (recognition and question-answering) with image generation data (mentioned in the Sec.~\ref{sec:image_generation}). However, our experiments reveal that naive joint training with mixed data leads to degraded performance compared to task-specific training. 

As shown in Fig.~\ref{fig:multi-tasks}, when tokens are trained exclusively for language generation, their application to image generation tasks results in outputs that fail to capture the target concepts (1st column). Conversely, tokens optimized solely for image generation prove inadequate for text generation tasks (2nd column). Furthermore, we find that joint training yields a compromised solution that underperforms in both domains, suggesting that the model struggles to learn representations that effectively serve both objectives simultaneously (3rd column). This observation aligns with previous work~\cite{hao2023toolkengpt,hard_prompt,visii} which suggests that tokens optimized for one specific task may lack semantic relevance for other tasks.

To overcome this limitation, we propose using two series of learnable tokens, each dedicated to a specific task. Specifically, the personalized concepts are represented as:
\begin{center}
    ``\texttt{\textcolor{blue}{<sks>} is \textcolor{red}{<$\text{g-tokens}$>}\textcolor{mygreen}{<$\text{u-tokens}$>}}.''
\end{center}
where \texttt{\textcolor{red}{<$\text{g-tokens}$>}} and \texttt{\textcolor{mygreen}{$<$$\text{u-tokens}$$>$}} represent $k$ and $h$ learnable tokens for image generation and understanding.

\begin{figure}[t]
    \centering
    \includegraphics[width=1\linewidth]{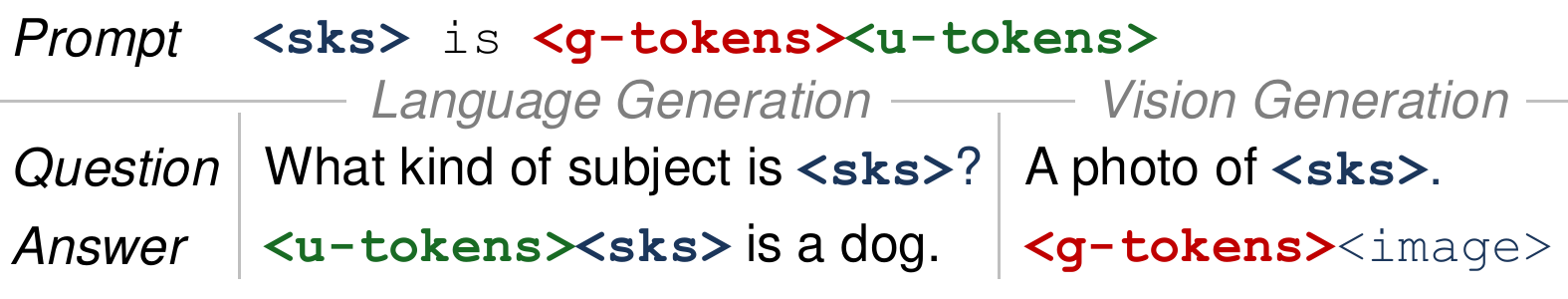}
    \caption{Self-prompting mechanism. When multiple tasks are presented, the model first predicts which information (latent tokens) should be used for this task first, and then performs the task.}
    \label{fig:selfprompting}
    \vspace{-3mm}
\end{figure}

During training, to force the model to learn the distinct roles of the two sets of tokens, we create the training data such that the model first predicts which set of tokens (\texttt{\textcolor{red}{<$\text{g-tokens}$>}} vs. \texttt{\textcolor{mygreen}{$<$$\text{u-tokens}$$>$}}) will be used for the task. We refer to this as ``self-prompting'' as the model needs to prompt itself first; Fig.~\ref{fig:selfprompting} shows examples. For instance, for text understanding tasks (e.g., ``What kind of object is \sks?''), the target output first includes \texttt{\textcolor{mygreen}{$<$$\text{u-tokens}$$>$}}, followed by the actual answer. The same technique is applied for image generation tasks. By requiring the model to first predict the appropriate token set, we force it to align the corresponding tokens with each task. 

This is partially inspired by \cite{hao2023toolkengpt}, where multiple tokens are used for calling different tools/tasks (e.g., mathematics, robot actions, etc). However, the key difference is that the task token in \cite{hao2023toolkengpt} is solely used for tool calling. In our approach, these tokens not only serve as task-mode calling tokens (i.e., for image or text generation) but also function as latent tokens, which contain the information needed to perform the task. This approach is flexible and could be adopted to other modalities as well (e.g., audio), and self-prompt tokens could be designed in a different way.  We leave these possibilities for future work.

\begin{figure*}[t]
  \centering
    \includegraphics[width=0.98\textwidth]{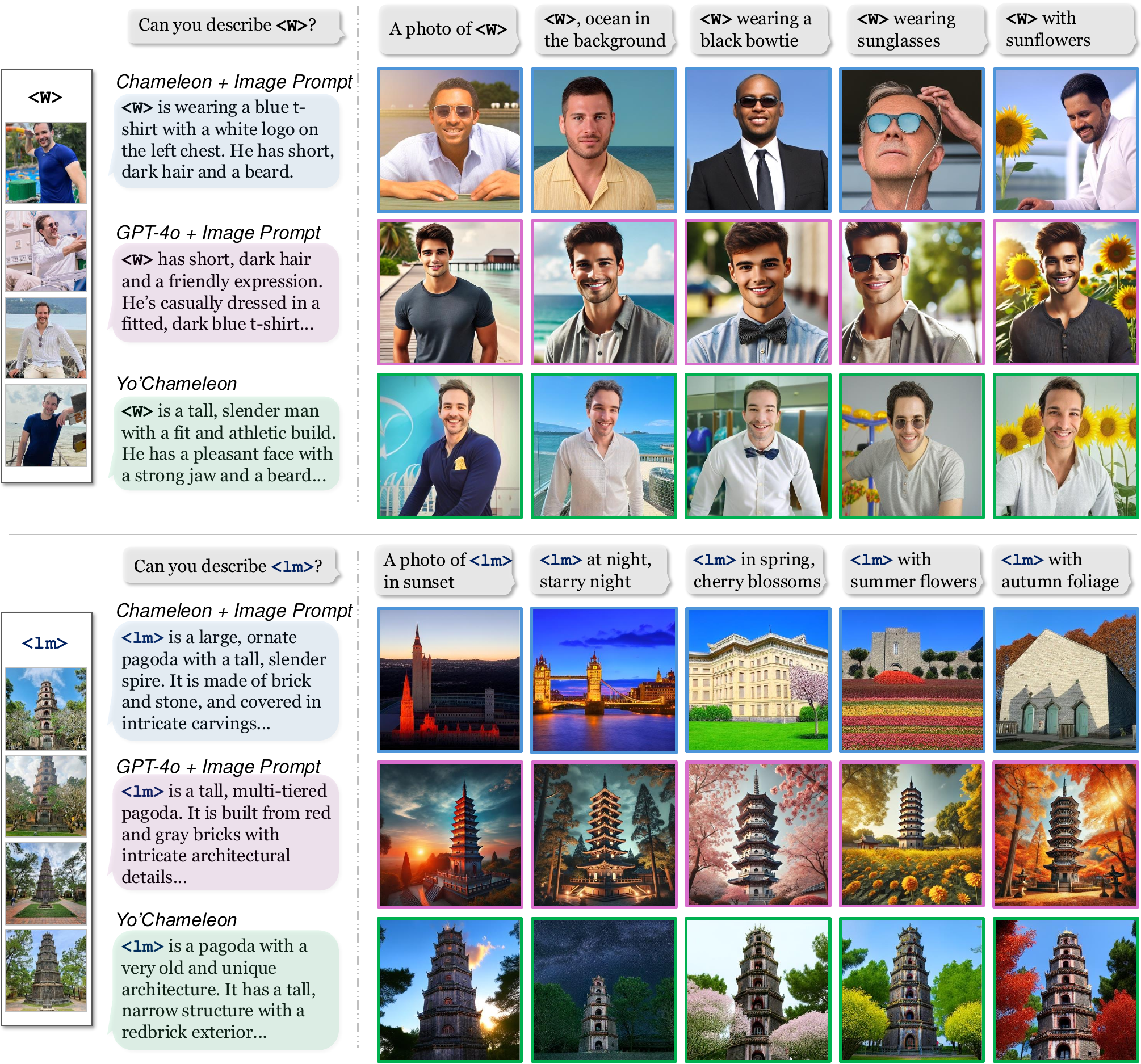}
    \caption{Qualitative comparison with Chameleon~\cite{chameleon} and GPT-4o~\cite{gpt4o} on image prompting. Yo'Chameleon (Ours) demonstrates more precise and personalized image generation.}
  \label{fig:qualitative}
  \vspace{-3mm}
\end{figure*}

\section{Experiments}
\label{sec:experiment}

\textbf{Training.} Unless otherwise stated, we use $n=4$ input images per concept and $k=h=16$ tokens to form a learnable prompt for each task, resulting in a total of $32$ latent tokens for personalized concepts. For optimization, we employ AdamW~\cite{adamw} with a learning rate of $1\times10^{-4}$. Each concept is trained for 15 epochs, with the best checkpoint selected based on a composite score averaging recognition accuracy and generation quality (measured by CLIP image similarity with training examples). All experiments are conducted on an A100 GPUs with a batch size of 4.

We choose Chameleon~\cite{chameleon} as our base model due to its simplicity in objective function (autoregressive for both text and image generation) and its unified LMM architecture. It is worth noting that our method generalizes to other LMMs, as it relies solely on token-level optimizations rather than model-specific architectures. While Chameleon was not originally published with image generation capabilities, we use the checkpoint from Anole~\cite{anole}, which recovered these capabilities through fine-tuning on an image generation dataset.

\textbf{Baselines.} The most straightforward baseline is using the base model (Chameleon~\cite{chameleon}) with personalized text and image prompting.
For personalized text prompting, we first obtain detailed captions of each concept by providing reference images to GPT-4o~\cite{gpt4o}.
These captions are then human-audited, and appended to Chameleon's system prompt (e.g., ``\texttt{<sks>} is a cinnamon-colored Shiba Inu with...''). For personalized image prompting, we append the reference image(s) (e.g., ``This is a photo of \texttt{<sks>}\texttt{<image>}"). Additionally, we compare our approach with GPT-4o~\cite{gpt4o}, a proprietary multimodal chatbot, using the same two types of personalized text and image prompts.

\textbf{Dataset.} We utilize the Yo'LLaVA dataset~\cite{yollava}, which consists of 40 subjects (10 humans and 30 non-human concepts). For negative images, we retrieve them from LAION-5B~\cite{laion5b} based on the average CLIP Image Similarity~\cite{clip} score between retrieved images and the mean feature representation of positive examples. After filtering NSFW content, we obtain approximately 1,000 negative images per concept. These images serve as ``soft-positive'' examples, with the top 100 most similar images designated as hard-examples for recognition. Additionally, we randomly sample 100 easy-negative examples from LAION-5B~\cite{laion5b}, which remain consistent across all concepts. In total, approximately 1,100 negative images are used for training each concept.

\textbf{Metrics.} To evaluate image understanding and text generation, we assess the model's recognition accuracy and question-answering ability. In total, there are 333 positive and 13,000 negative images for recognition. During testing, we present a photo and ask the model ``Is \texttt{<sks>} in this photo?" The ground-truth answer is either ``Yes" or ``No". We use a weighted accuracy metric to balance the positive and negative classes, following the protocol in~\cite{yollava}. For question-answering, we provide multiple-choice questions (A or B) with 100 visual and 400 text-based questions.

For image generation, we produce 100 images per concept using the prompt ``A photo of \texttt{<sks>}" and compute the CLIP Image Similarity Score~\cite{clip} between the generated images and positive examples. Additionally, in ablation studies, we further extend our analysis by reporting the Facial Similarity Score between the generated and positive images for 10 human faces (where applicable) using the off-the-shelf facial feature extractor ArcFace~\cite{deng2018arcface}.

%
%
%
%

\begin{figure*}
\begin{minipage}{\textwidth}
  \begin{minipage}{0.6\textwidth} 
    \centering
    \scalebox{0.7}{ 
    \setlength{\tabcolsep}{4.5pt} 
    \begin{tabular}{l|ccccc|ccc}
    \toprule
     & Ours & Chameleon & \multicolumn{3}{c|}{Chameleon~\cite{chameleon} + Prompt} & \multicolumn{3}{c}{GPT-4o~\cite{gpt4o} + Prompt} \\
    \cmidrule(lr){2-2} \cmidrule(lr){3-3} \cmidrule(lr){4-6} \cmidrule(lr){7-9}
    Type & Learnable & $\emptyset$ & Text & \multicolumn{2}{c|}{Image} & Text & \multicolumn{2}{c}{Image} \\
    \cmidrule(lr){2-2} \cmidrule(lr){3-3} \cmidrule(lr){4-4} \cmidrule(lr){5-6} \cmidrule(lr){7-7} \cmidrule(lr){8-9}
    \# tokens & 32 & 0 & $\sim$64 & $\sim$1k & $\sim$4k & $\sim$64 & $\sim$1k & $\sim$4k \\
    \midrule
    Recognition Accuracy & \textbf{0.845} & 0.500 & \underline{0.727} & 0.361 & 0.327 & 0.841 & 0.902 & 0.915 \\
    \midrule
    \multicolumn{9}{l}{Question Answering} \\
    \quad Visual & \textbf{0.604} & 0.474 & 0.523 & \underline{0.580} & 0.547 & 0.923 & 0.867 & 0.887 \\
    \quad Text & \textbf{0.721} & 0.405 & \underline{0.716} & 0.573 & 0.231 & 0.798 & 0.982 & 0.978\\
    \midrule
    \multicolumn{9}{l}{Image Generation} \\
    \quad CLIP-I & \textbf{0.783} & 0.425 & 0.566 & 0.487 & \underline{0.589} & 0.636 & 0.657$^{\star}$ & 0.680$^{\star}$ \\
    \quad Facial Sim & \textbf{0.212} & 0.009 & 0.012 & 0.013 & \underline{0.059} & 0.028 & 0.036$^{\star}$ & 0.063$^{\star}$ \\
    \bottomrule
    \end{tabular}
    }
    \captionsetup{hypcap=false}
    \captionof{table}{Comparisons with Chameleon~\cite{chameleon} and GPT-4o~\cite{gpt4o} using personalized image/text prompts. Our approach achieves significantly improved personalized image generation capabilities.}
    \label{tab:baseline}
  \end{minipage}
  \hfill
  \begin{minipage}{0.38\textwidth}
    \centering
    \vspace{-3mm}
    \includegraphics[width=1\textwidth]{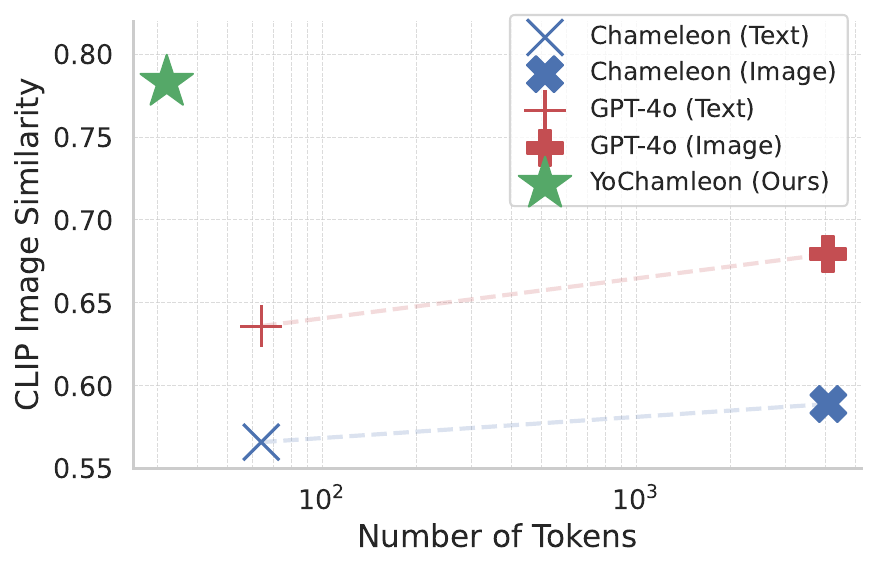}
    \vspace{-7mm}
    \captionsetup{hypcap=false}
    \captionof{figure}{Number of Prompting Tokens vs. Personalized Image Generation Quality.}
    \label{fig:token_vs_acc}
  \end{minipage}
\vspace{-1mm}
\end{minipage}
\end{figure*}


\subsection{Personalized Language Generation}

Tab.~\ref{tab:baseline} shows the recognition and question-answering abilities of the evaluated models. The vanilla Chameleon model, lacking personalized concept information, performs essentially at random (0.474–0.500) on both tasks. With the addition of personalized text prompts, Chameleon's performance improves (0.523–0.727).
Image prompting shows mixed results. When given a single example ($\sim$1k column), it improves question-answering but does not enhance recognition accuracy. Providing multiple images ($\sim$4k column) generally leads to a drop in performance on both tasks.

Notably, our approach outperforms Chameleon across all language generation tasks, with the recognition accuracy increases significantly (0.727 to 0.845). We achieve these improvements using fewer tokens (32) compared to the detailed text ($\sim$64) or image ($\sim$1k) prompting of Chameleon.

We also present GPT-4o's results for reference. GPT-4o performs well with both text and image prompts. For recognition tasks, our approach achieves comparable results (0.845 vs.\ 0.902) while requiring significantly fewer tokens (32 vs.\ $\sim$1k). For question answering, GPT-4o demonstrates better performance. This discrepancy can be attributed to two factors: (1) our use of a less powerful base model (i.e., Chameleon), and (2) the question data from~\cite{yollava} being relatively simple and generic (e.g., ``What is the color of this subject?", ``What material is this subject made of?"), where text descriptions as prompt are often sufficient. This explains why we achieve comparable results in recognition tasks, which require more fine-grained visual details. Therefore, we believe our approach offers value in terms of token efficiency while maintaining competitive performance.


\subsection{Personalized Image Generation}

Personalized image generation is generally a more challenging task than language generation. This is because recreating novel concepts with pixel-level detail is much more complex than simply answering questions based on existing references.
In these cases, our learnable prompts with Chameleon clearly show advantages, outperforming all other methods by a significant margin.  Specifically, Tab.~\ref{tab:baseline} clearly shows that Yo'Chameleon achieves the highest CLIP Image Similarity Score (0.783), significantly surpassing the scores for Chameleon with either Image/Text Prompts (0.566–0.487).

When compared with GPT-4o~\cite{gpt4o}, we find that GPT-4o generally captures high-level semantic details of personalized concepts reasonably well with both image and text prompts (i.e., 0.636–0.657). However, it struggles to capture the nuanced details of personalized subjects (see Fig.~\ref{fig:qualitative}).
This limitation is evident in the Facial Similarity Score, where we compare generated images to real images of 10 human faces.
GPT-4o’s generated images show low similarity to the actual person (e.g., 0.028–0.036), while Yo'Chameleon generated images more accurately capture facial details, making it far more suitable for personalization.

\begin{figure*}[t]
  \centering
    \includegraphics[width=1\linewidth]{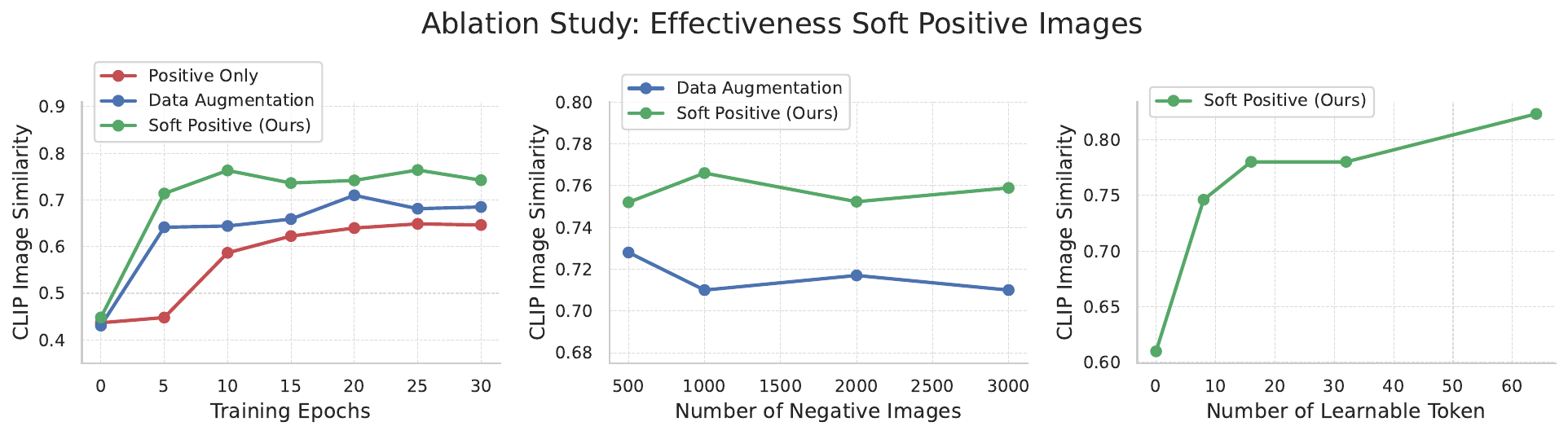}
    \vspace{-6mm}
    \caption{Ablation studies for image generation tasks. Overall, using ``soft positive'' and increasing the latent tokens boost performance.}
    \vspace{-1mm}
  \label{fig:ablation}
  
\end{figure*}

\section{Ablation Studies}
\label{sec:ablation}

In all subsequent studies, we use 10 human faces from the Yo'LLaVA~\cite{yollava} dataset, and evaluate the Facial Similarity Score between the generated and positive images using ArcFace~\cite{deng2018arcface}, which is specifically trained to distinguish nuanced differences between faces, providing a reliable metric for personalization generation.  We ablate the: (1)  importance of ``soft positive'' images, (2) number of ``soft positive" images, (3) number of learnable tokens for the image generation task, and (4) different training strategies. As the focus of the first three experiments is on image generation, we train only with image generation data for these experiments, while the last one is trained with a mixture of data. The number of trainable tokens for each task are set to $k=16$.

\textbf{Importance of ``soft positive'' images.} We compare our gradually added negative images with three main baselines: (1) Positive only (2–3 images), (2) Data augmentation via inpainting (1000 images), and (3) Soft Positive Images (Ours). For data augmentation, we first use SAM~\cite{kirillov2023segment} to extract the foreground of the subjects, then randomly resize the subject to 30–70\% of the 512x512 image size, and inpaint the background using Stable Diffusion-XL~\cite{podell2023sdxlimprovinglatentdiffusion}. We generate 100 background captions with GPT-4o~\cite{gpt4o}, which are then human-audited. Results are presented in Fig.~\ref{fig:ablation} (first plot). As shown, while data augmentation improves the results compared to training with positive images only, it still falls short of the performance achieved by ``soft positive'' images.

\textbf{Number of augmented/ ``soft-positive'' images.} We next investigate the impact of varying the number of soft positive images (including both hard-negative and inpainting-augmented samples) used during training. Results demonstrate that using soft positive images consistently outperforms augmented images (Fig.~\ref{fig:ablation}, second plot). This superiority likely stems from the inherent limitations of segmentation and inpainting models for augmented data. 


\textbf{Number of learnable tokens.} With the number of ``soft-positive" images fixed at 1,000, we vary the number of trainable tokens $k$ from 0 to 64. $k=0$ means no latent tokens are trained for this task. Overall, increasing the number of trainable tokens improves the quality of image generation (Fig.~\ref{fig:ablation}, third plot). Quantitatively and qualitatively, we find that 16 tokens achieve reasonable results for most concepts, making it an effective and compact choice. For higher-quality image generation, one may increase the number of latent tokens. Empirically, we also note that, although the generated subjects appear visually similar, there is still room for improvement in the accuracy of generated human faces (e.g., current facial similarity is 0.212, while the threshold for a good human facial similarity would be 0.4 or higher).

\textbf{Different training strategies.} Our approach employs separate tokens for each task (understanding and generation) with self-prompting prior to prediction. We next validate this design.
We begin by training the same set of tokens for two different tasks (Shared learnable prompt, in Table~\ref{tab:training_strategy}). Results indicate that using a single set of tokens and training them specifically for each task achieves optimal performance for that particular task. For example, (1) Language data only: achieves the best recognition accuracy but fails to generate images. For (2) Image generation data only, we explore three variations: (2.1) Positive only: training exclusively with positive images; (2.2) Negative + Finetune: treating all negative samples equally during training across all tokens, followed by fine-tuning with positive images; and (3) Soft-positive (Ours): gradually incorporating more tokens as soft-positive images become more similar to true positives. Our results demonstrate that the soft-positive strategy achieves the best results. (3) Training on mixture data yields intermediate scores across both tasks, suggesting that generation and understanding tasks may not trivially be complementary.

The above findings suggest two key insights: (1) our proposed approach of adaptively setting soft-positive images with varying token lengths is more effective for generation tasks, and (2) shared learnable prompts are suboptimal when handling multiple tasks simultaneously, necessitating separate token sets for different tasks.

For the separate learnable prompts approach, we also ablate different strategies: (1) Concatenate: training two sets of tokens independently for each task and concatenating them at test time; (2) Concatenate + Fine-tune: extending strategy (1) with an additional fine-tuning step post-concatenation; and (3) Self-prompting (Ours): our proposed mechanism that first predicts prompt tokens before making the actual prediction. Results demonstrate that our self-prompting approach achieves optimal performance, matching the effectiveness of task-specific token training.

\begin{table}[t]
\small
  \centering
\resizebox{1\columnwidth}{!}{ 
  \begin{tabular}[t]{lccc}
    \toprule
     & Acc. ($\uparrow$) & CLIP-I ($\uparrow$) & Face Sim ($\uparrow$)\\
    \midrule
    \multicolumn{3}{l}{\textbf{Shared learnable prompt} \textit{(16 tokens in total)}} \\
    Language data only & \textbf{0.784} & 0.120 & 0.032\\
    Image generation data only & & & \\
    \quad Positive only & 0.104 & 0.678 & 0.188 \\
    \quad Negative + Finetune & 0.004 & 0.711 & 0.193\\
    \quad Soft positive & 0.108 & \underline{0.742} & \textbf{0.225} \\
    Mixture data  & 0.564 & 0.687 & 0.193\\
    Mixture data (32 tokens) & 0.562 & 0.684 & 0.194 \\
    \midrule
    \multicolumn{4}{l}{\textbf{Separated learnable prompt} \textit{(32 tokens in total)}} \\
    Concatenate & 0.502 & 0.615 & 0.156 \\
    Concatenate + Finetune & 0.251 & 0.648 & 0.189\\
    Self-Prompting (Ours) & \underline{0.747} & \textbf{0.761} & \underline{0.224} \\
    \bottomrule
  \end{tabular}
  }
  \vskip -0.1in
  \caption{Ablation studies on different training strategy. We use recognition accuracy to evaluate understanding capability, and CLIP and Face similarities for image generation quality. }
    \label{tab:training_strategy}
    \vspace{-2mm}
\end{table}
\section{Conclusion}
\label{sec:conclusion}


We presented the first attempt to personalize Large Multimodal Models (LMMs) for both vision and language understanding and generation tasks. We introduced a dual prefix prompt architecture with a self-prompting mechanism to achieve strong performance in both understanding and generation capabilities. We also proposed a novel soft-positive training strategy that leverages hard-negative samples to enhance generation quality in spite of limited user data. Experimental results demonstrated that our approach successfully maintains the model's general knowledge while enabling effective personalization across both tasks, representing a significant step toward making LMMs more personally relevant for real-world applications.

\section*{Acknowledgment}
This work was supported in part by NSF IIS2404180, Adobe Data Science award, Microsoft Accelerate Foundation Models Research Program, and Institute of Information \& communications Technology Planning \& Evaluation (IITP) grants funded by the Korea government (MSIT) (No. 2022-0-00871, Development of AI Autonomy and Knowledge Enhancement for AI Agent Collaboration) and (No. RS-2022-00187238, Development of Large Korean Language Model Technology for Efficient Pre-training).
{
    \small
    \bibliographystyle{unsrt}
    \bibliography{main}
}

%
%
%
%
\clearpage
\setcounter{page}{1}
\maketitlesupplementary


\section{Full-model Finetuning vs. Soft Prompt}
\label{sec:forgetting}
As discussed in the Introduction, our experiment reveals that soft prompt tuning can match the performance of full-model fine-tuning when trained with approximately 300 real images of a single concept. In this section, we provide details about that experiments.

In this experiment, we collected photos for three concepts: one person (300 images), one dog (500 images), and one cat (500 images). These images are ``in-the-wild'' and therefore exhibit significant diversity in appearance. To address this, we first roughly cropped the regions containing the target concepts, creating datasets for each concept at a resolution of \(512 \times 512\). The concepts of interest are typically centered within the images.
Our goal was to verify whether soft prompt tuning could achieve performance comparable to full-model fine-tuning, which is commonly used in personalized image generation (i.e.,~\cite{ruiz2023dreambooth,instantbooth,textual_inversion}).

For full-model fine-tuning, we fine-tune Chameleon~\cite{chameleon} using the prompt ``A photo of \texttt{<sks>}'' with a learning rate of \(1 \times 10^{-7}\), a batch size of 2, over maximum 1000 iterations. For soft prompt tuning, we used the prompt ``\texttt{<sks>} is \texttt{<token$_{1}$>}$...$\texttt{<token$_{16}$>}. A photo of \texttt{<sks>}.'' with a learning rate of \(1 \times 10^{-4}\), batch size of 4, for 15 epochs. In another words, a concept is represented by $k=16$ latent tokens.

To evaluate general abilities, we used prominent benchmarks such as MMLU~\cite{mmlu} for text-only generation, POPE~\cite{pope}, and MMBench~\cite{MMBench} for visual question answering. For personalized abilities, we measured CLIP-Image Similarity~\cite{clip} and Facial Similarity using the off-the-shelf ArcFace model~\cite{deng2018arcface} to compare generated images with the reference images.

\begin{figure}[ht]
    \centering
    \includegraphics[width=0.99\linewidth]{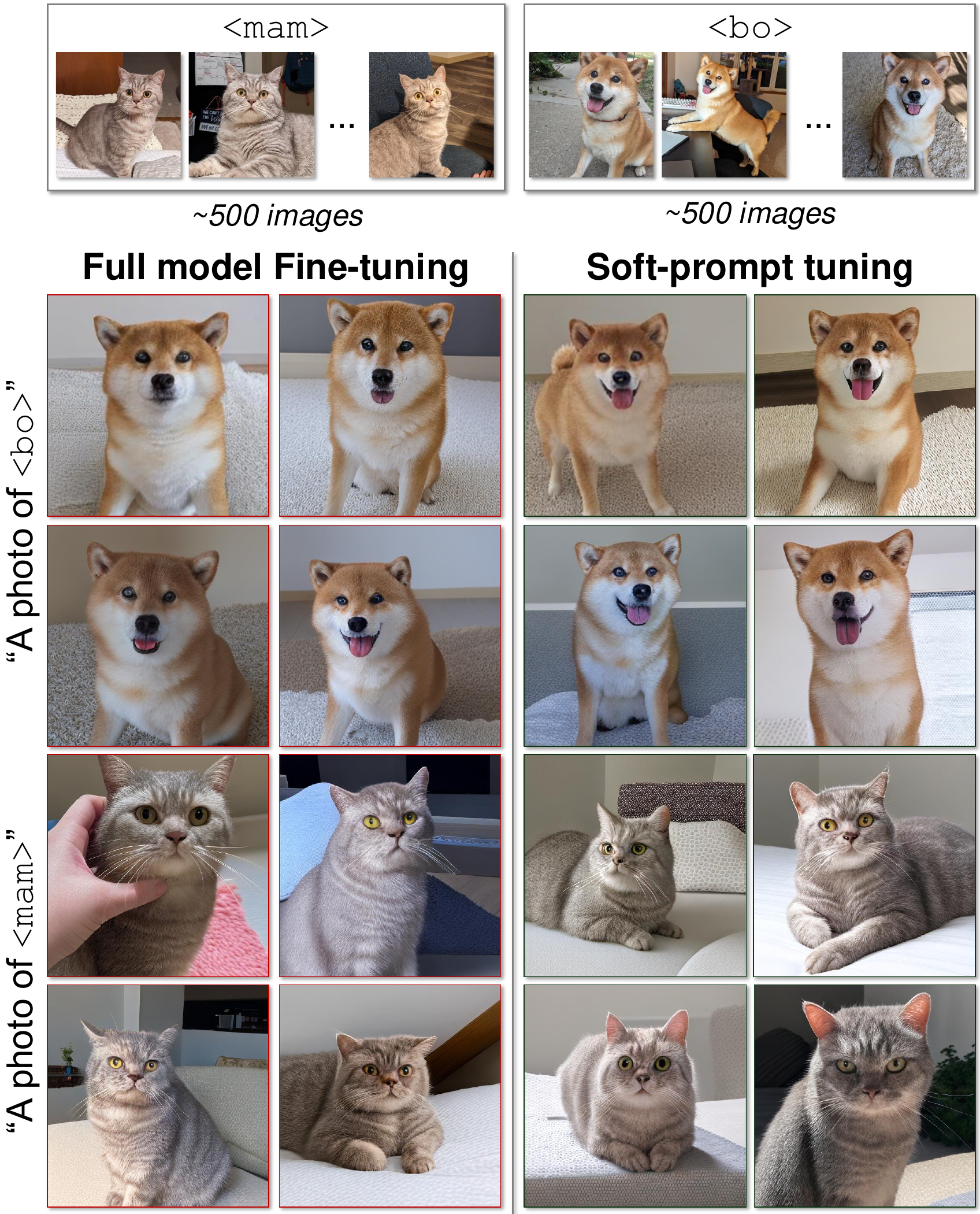}
    \caption{With 300+ real images, soft-prompt tuning can match the performance of full-model fine-tuning while retaining the model's overall abilities. We cannot show the facial results due to anonymity. }
    \label{fig:fullmodel-finetune}
\end{figure}

The results are shown in Tab.~\ref{tab:fullmodel-finetune} (first five rows, ``300+ real images''). As the table demonstrates, full-model fine-tuning leads to catastrophic forgetting, with performance degradation ranging from 1--65\% across tasks. Although fine-tuning improves personalized image generation metrics (e.g., CLIP-Image Similarity increases from 0.804 to 0.849), it significantly compromises the model's general abilities, such as text-only generation, where MMLU performance drops from 65.4 to 59.6. In contrast, soft prompt tuning achieves comparable performance in personalized image generation (e.g., Facial Similarity reaches 0.429) while maintaining general abilities nearly identical to the base model.

It is important to note that this experiment was conducted for research purposes only and has limited practical applicability, as users might not be able, or not willing to provide 300+ images of a concept. Nonetheless, this pilot study effectively demonstrates the advantages of soft prompt tuning over full-model fine-tuning: (1) it matches the performance of full-model fine-tuning in personalized tasks and (2) mitigates catastrophic forgetting.

\begin{figure*}[ht]
\begin{minipage}{\textwidth}
  \begin{minipage}{1\textwidth} 
    \centering
    \scalebox{0.82}{ 
    \setlength{\tabcolsep}{4.5pt} 
    \begin{tabular}{l|ccccccccccc}
    \toprule
     & \multicolumn{5}{c}{General Abilities ($\uparrow$)} & \multicolumn{2}{c}{Personalized Image Gen. ($\uparrow$)} \\
    \cmidrule(lr){2-6} \cmidrule(lr){7-8}
     & \multicolumn{3}{c}{POPE~\cite{pope}} & MMBench~\cite{MMBench}& \multirow{2}{*}{MMLU~\cite{mmlu}} &\multirow{2}{*}{CLIP-I~\cite{clip}} & \multirow{2}{*}{Facial Sim~\cite{deng2018arcface}} \\
    \cmidrule(lr){2-4} \cmidrule(lr){5-5}
    Settings & pop & rand & adv & en &  &  &  \\
    \midrule
    \textit{Random} & \textit{0.500} & \textit{0.500} & \textit{0.500}  & \textit{0.25} & \textit{0.25}  & $\sim$0.3-0.5 & $\sim$ 0.001 \\
    
    Original Chameleon~\cite{chameleon} & 0.702 & 0.504 & 0.656  & 0.57 & 0.52  & 0.423 & 0.001\\
    \midrule
    \textbf{300+ real images} (3 concepts)\\
    
    \quad Soft Prompt (16 tokens) & 0.702 {\raisebox{0.1ex}{\scriptsize\textcolor{mygreen}{(same)}}} & 0.504 {\raisebox{0.1ex}{\scriptsize\textcolor{mygreen}{(same)}}} & 0.656 {\raisebox{0.1ex}{\scriptsize\textcolor{mygreen}{(same)}}} & 0.57 {\raisebox{0.1ex}{\scriptsize\textcolor{mygreen}{(same)}}} & 0.50 {\raisebox{0.1ex}{\scriptsize\textcolor{red}{($-$3.8 \%)}}} & 0.803 {\raisebox{0.1ex}{\scriptsize\textcolor{mygreen}{($+$0.380)}}} & 0.427 {\raisebox{0.1ex}{\scriptsize\textcolor{mygreen}{($+$0.426)}}}\\
    
    \quad Full-model Fine-tuning (iter=300) & 0.561 {\raisebox{0.1ex}{\scriptsize\textcolor{red}{($-$20.1\%)}}} & 0.497 {\raisebox{0.1ex}{\scriptsize\textcolor{red}{($-$1.4\%)}}} & 0.534 {\raisebox{0.1ex}{\scriptsize\textcolor{red}{($-$18.6\%)}}} & 0.46 {\raisebox{0.1ex}{\scriptsize\textcolor{red}{($-$19.3\%)}}}& 0.21 {\raisebox{0.1ex}{\scriptsize\textcolor{red}{($-$59.6\%)}}} & 0.804 {\raisebox{0.1ex}{\scriptsize\textcolor{mygreen}{($+$0.381)}}}& 0.429 {\raisebox{0.1ex}{\scriptsize\textcolor{mygreen}{($+$0.428)}}}\\
    
    \quad Full-model Fine-tuning (iter=500) & 0.500 {\raisebox{0.1ex}{\scriptsize\textcolor{red}{($-$28.8\%)}}} & 0.500 {\raisebox{0.1ex}{\scriptsize\textcolor{red}{($-$0.8\%)}}} & 0.500 {\raisebox{0.1ex}{\scriptsize\textcolor{red}{($-$23.8\%)}}} & 0.45 {\raisebox{0.1ex}{\scriptsize\textcolor{red}{($-$21.1\%)}}}& 0.18 {\raisebox{0.1ex}{\scriptsize\textcolor{red}{($-$65.4\%)}}} &  0.849 {\raisebox{0.1ex}{\scriptsize\textcolor{mygreen}{($+$0.426)}}} & 0.429 {\raisebox{0.1ex}{\scriptsize\textcolor{mygreen}{($+$0.428)}}}\\
    
    \midrule
    \textbf{3-5 images} (10 concepts)\\
    
    \quad Soft Prompt (16 tokens) & 0.702 {\raisebox{0.1ex}{\scriptsize\textcolor{mygreen}{(same)}}} & 0.504 {\raisebox{0.1ex}{\scriptsize\textcolor{mygreen}{(same)}}} & 0.656 {\raisebox{0.1ex}{\scriptsize\textcolor{mygreen}{(same)}}} & 0.57 {\raisebox{0.1ex}{\scriptsize\textcolor{mygreen}{(same)}}} & 0.51 {\raisebox{0.1ex}{\scriptsize\textcolor{red}{($-$1.9\%)}}}&  0.742 {\raisebox{0.1ex}{\scriptsize\textcolor{mygreen}{($+$0.319)}}} & 0.225 {\raisebox{0.1ex}{\scriptsize\textcolor{mygreen}{($+$0.224)}}}\\
    
    \quad Full-model Fine-tuning (iter=300) & 0.500 {\raisebox{0.1ex}{\scriptsize\textcolor{red}{($-$28.8\%)}}} & 0.500 {\raisebox{0.1ex}{\scriptsize\textcolor{red}{($-$0.8\%)}}} & 0.500 {\raisebox{0.1ex}{\scriptsize\textcolor{red}{($-$23.8\%)}}} & 0.45 {\raisebox{0.1ex}{\scriptsize\textcolor{red}{($-$21.1\%)}}} & 0.20 {\raisebox{0.1ex}{\scriptsize\textcolor{red}{($-$61.5\%)}}}&  0.748 {\raisebox{0.1ex}{\scriptsize\textcolor{mygreen}{($+$0.325)}}}& 0.242 {\raisebox{0.1ex}{\scriptsize\textcolor{mygreen}{($+$0.241)}}} \\
    \bottomrule
    \end{tabular}
    }
    \captionsetup{hypcap=false}
    \captionof{table}{Soft-Prompt Tuning vs. Full-Model Fine-Tuning. Overall, soft-prompt tuning matches the performance of full-model fine-tuning for personalized abilities while retaining the original model's general capabilities.
    }
    \label{tab:fullmodel-finetune}
  \end{minipage}
  \hfill
\vspace{-1mm}
\end{minipage}
\end{figure*}

\section{Additional Ablation Studies}
Along with the ablation studies presented in the main paper, we provide an additional ablation study on (1) the number of ``soft-positive'' images and (2) Evaluation for Catastrophic Forgetting . These studies could not be included in the main paper due to space limitations.

\subsection{Number of Soft-Positive Images}
\label{sec:qualitative_for_ablation_studies}
Similar to other ablation studiesin the main paper, this study aims to analyze the effect of varying the number of ``soft-positive'' images used during concept training. We vary the number of ``soft-positive'' images from 0 to 1000, where 0 indicates no ``soft-positive'' images were used during training, and 1000 indicates that 1000 ``soft-positive'' images were included.

\begin{figure}
    \centering
    \includegraphics[width=0.99\linewidth]{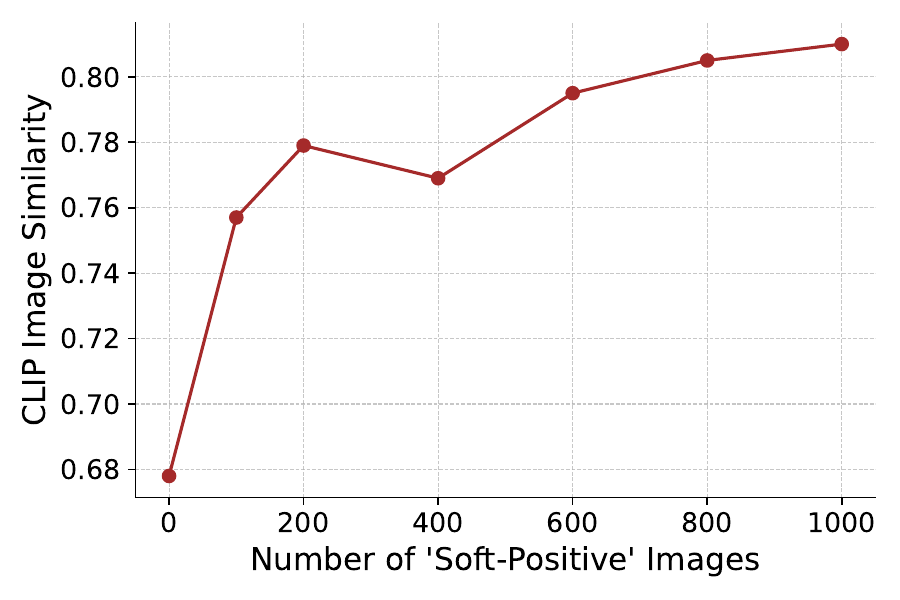}
    \caption{Ablation studies on the number of ``soft-positive'' images. Generally, increasing the number of ``soft-positive'' images helps to boost performance.}
    \label{fig:number-of-soft-positive}
    \vspace{-3mm}
\end{figure}

The results are presented in Fig.~\ref{fig:number-of-soft-positive}. As shown, incorporating ``soft-positive'' images significantly improves performance compared to training with only positive images (e.g., 0.68 vs. 0.76+). Overall, increasing the number of ``soft-positive'' images enhances performance, with saturation observed around 1000 images when training with a soft prompt of token length $k=16$ tokens.

\subsection{Catastrophic Forgetting}
Similar to the evaluation in Sec.~\ref{sec:forgetting}, for general abilities, we utilized prominent benchmarks such as MMLU~\cite{mmlu} for text-only generation, POPE~\cite{pope}, and MMBench~\cite{MMBench} for visual question answering. For personalized abilities, we evaluated CLIP-Image Similarity~\cite{clip} and Facial Similarity using the off-the-shelf ArcFace model~\cite{deng2018arcface} to compare generated images with the reference images.

The results are presented in Tab.~\ref{tab:fullmodel-finetune} (last three rows). As shown, full-model finetuning leads to catastrophic forgetting across all benchmarks, with performance drops ranging from 1\% to 61.5\%. In contrast, using soft prompts preserves the model's general performance across nearly all benchmarks while achieving personalized abilities comparable to full-model finetuning (e.g., CLIP-Image Similarity is 0.742 vs. 0.748).

\section{Data Augmentation Details}
\label{sec:qualitative_for_ablation_studies}
Here, we provide details about the data augmentation process for the ablation studies in the main paper. There are two main approaches for creating augmented training data: (A) Using positive images only, and (B) Using ``Soft-Positive'' Images (Ours).

\begin{figure}
    \centering
    \includegraphics[width=0.95\linewidth]{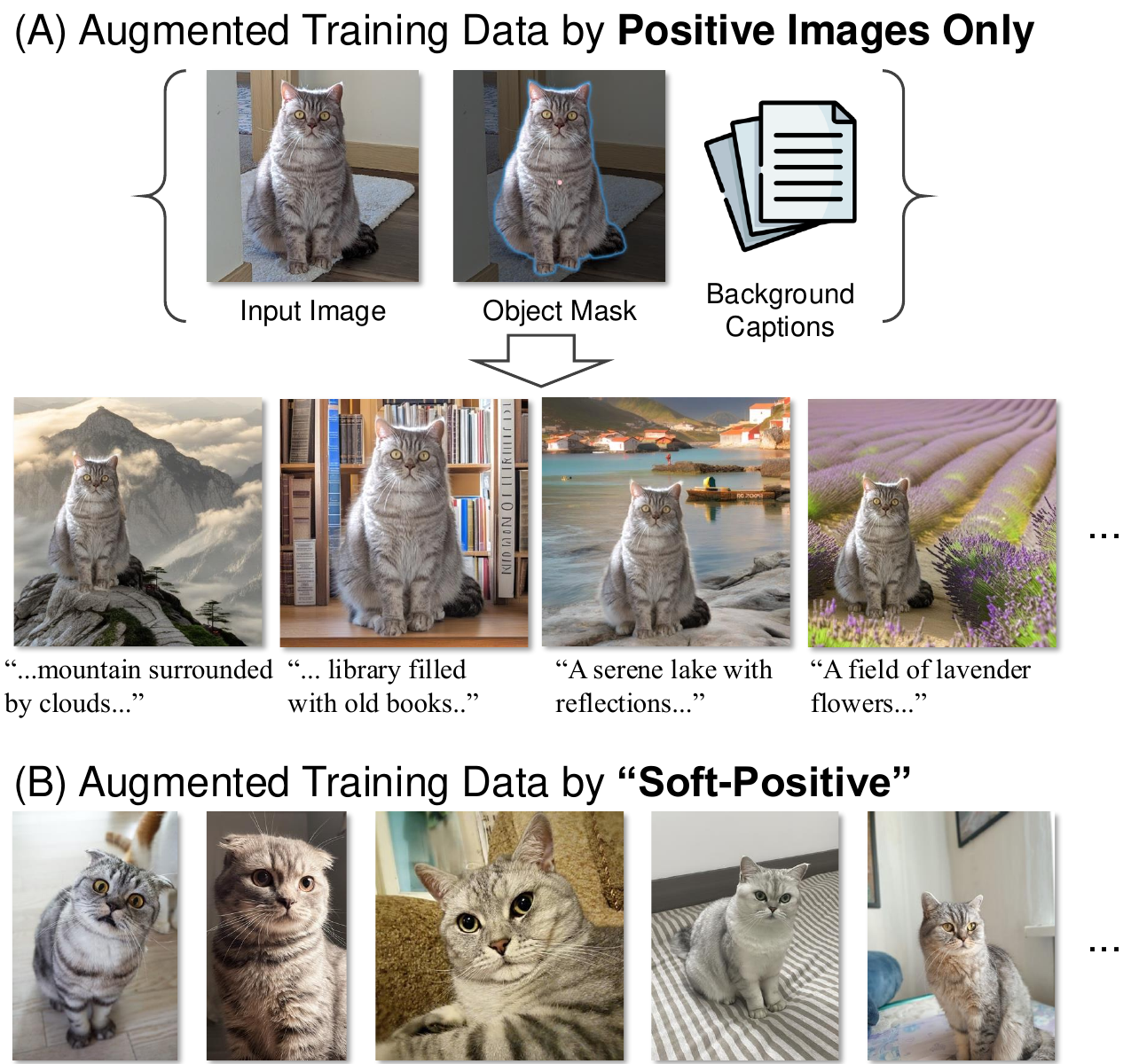}
    \caption{Comparison of data augmentation methods. Using `Soft-Positive'' images can increase both diversity and realism of the training data.}
    \label{fig:data-aug}
    \vspace{-3mm}
\end{figure}

\textbf{Augmentation with Positive Images Only.} The objective of this approach is to increase the diversity of training data when only 3--5 images of a subject are available. Inspired by~\cite{anythinganywhere}, given an input image (e.g., a photo of a cat), we first obtain the corresponding object mask (e.g., the segmentation mask of the cat) using a pretrained SAM~\cite{kirillov2023segment}. Subsequently, we randomly resize the subject (ranging from 30--100\%) within a $512 \times 512$ image. This resized subject is then paired with a randomly selected background caption from a background library to inpaint the background (e.g., ``A field of lavender flowers'') using StableDiffusion-XL~\cite{podell2023sdxlimprovinglatentdiffusion}. Fig.~\ref{fig:data-aug}A illustrates this process.

The background library contains 100 captions, all generated by GPT-4o~\cite{gpt4o} and later human-audited. Table~\ref{tab:background-caption} lists 10 randomly selected examples of these captions. All augmented images generated through this process are treated as positive images and are given equal weight as positive samples during training.

\textbf{Augmentation with ``Soft-Positive'' Images (Ours).}
In this approach, input images are used to retrieve the top $N$ most similar images from LAION-5B~\cite{laion5b}. These retrieved images are referred to as ``soft-positive'' images. The retrieved images are ranked, and an adaptive prompt length strategy is applied to describe them: the more similar a soft-positive image is to the input image, the more tokens are allocated to describe it. Fig.~\ref{fig:data-aug}B provides some examples of these ``soft-positive'' images.

\textbf{Comparisons.} A key limitation of (A) Augmentation with Positive Images Only is that while the backgrounds vary, the foreground subject remains the same, which might restrict diversity in terms of the subject's pose or other variations. In contrast, the ``soft-positive'' images not only provide diverse background information but also add variations in the foreground, such as pose and angle. 

Additionally, it is important to note that augmented images are generated content, whereas ``soft-positive'' images are real images. Training on real distributions can lead to more realistic results compared to training on generated (synthetic) distributions.

\section{Limitation}
\label{sec:limitation}
\begin{figure}[ht]
    \centering
    \includegraphics[width=1\linewidth]{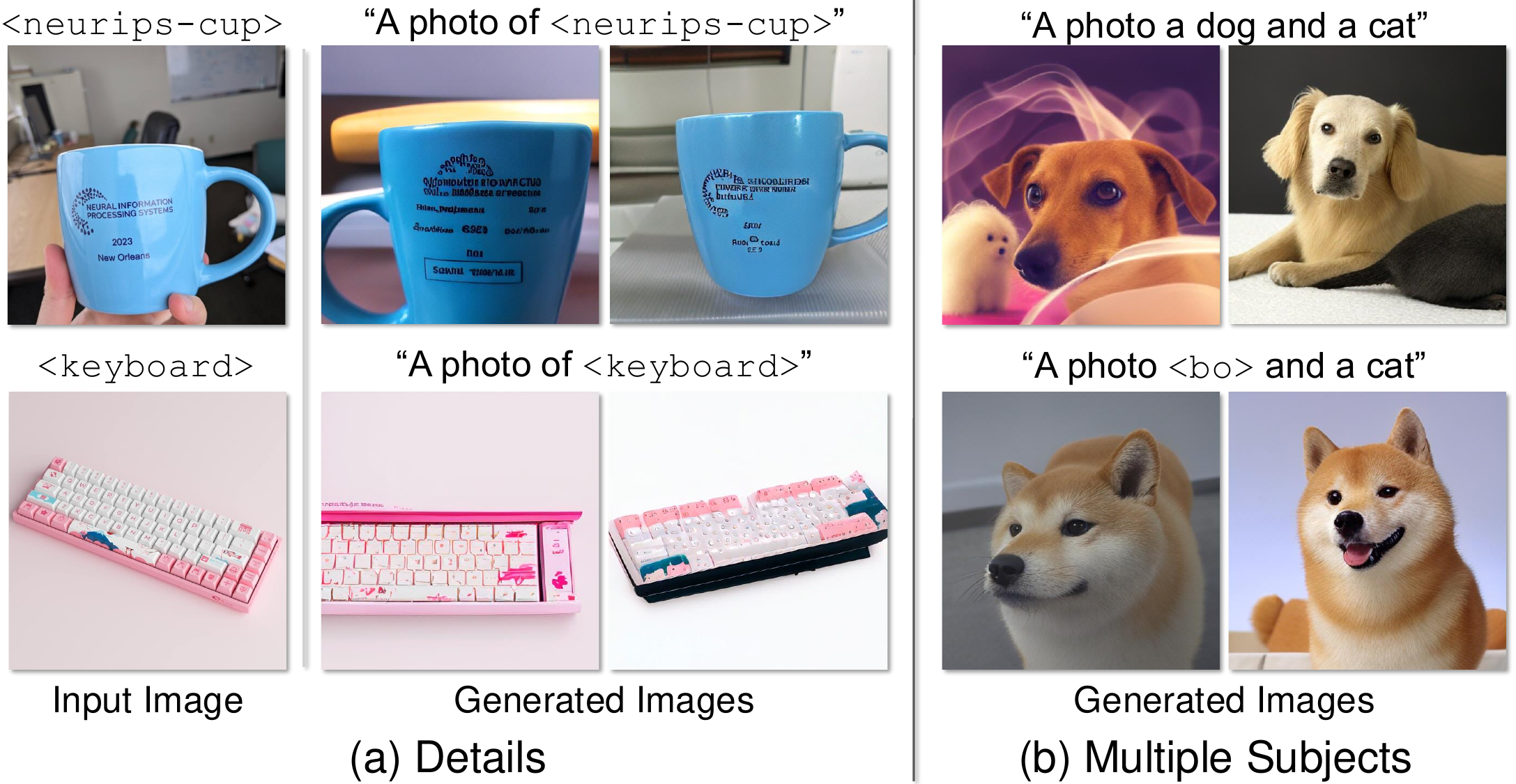}
    \caption{Limitations. (a) lacks of details; (b) Generate multiple subjects}
    \label{fig:limitation}
    \vspace{-3mm}
\end{figure}
Our method is not without limitations. The first limitation arises when dealing with objects that have intricate details (e.g., text on a cup or characters on a keyboard). Examples of such cases are shown in Fig.~\ref{fig:limitation}(a).

The second limitation is that, like other personalization methods~\cite{yollava,ruiz2023dreambooth,textual_inversion}, our method's performance is constrained by the capabilities of the base model. For instance, as \cite{yollava} highlights, personalized Vision-Language Models like LLaVA~\cite{liu2023llava} can still produce hallucinations (e.g., providing an incorrect date of birth for a person when asked). Similarly, our approach inherits the limitations of its underlying models, in this case, Chameleon/Anole~\cite{chameleon,anole}. While these models perform reasonably well in generating object-centric images (e.g., ``A photo of a dog''), they struggle with generating images involving multiple concepts (e.g., ``A photo of a dog and a cat,'' as shown in Fig.~\ref{fig:limitation}(b)). Consequently, we were unable to test our approach on multiple personalized concepts effectively.

Lastly, although we achieved encouraging results in personalizing for individuals (e.g., facial similarity of 0.2xx), there remains a significant gap when it comes to personalizing human faces. For reference, the recommended threshold for facial recognition similarity is around 0.4--0.5, highlighting considerable room for improvement in this area.


\begin{table}\centering
\begin{minipage}{0.99\columnwidth}\vspace{0mm}    \centering
\begin{tcolorbox}[boxrule=0.5pt]
    \centering
    \small
     \hspace{-6mm}
\begin{itemize}[leftmargin=0.1mm]
\setlength{\itemsep}{2pt}
    \item A serene beach with golden sand and clear blue water.
    \item A vibrant sunset over a calm ocean.
    \item A snowy village during a peaceful winter evening.
    \item A quiet library filled with old books and wooden shelves.
    \item A crowded street in an ancient Asian market.
    \item A colorful spring garden in full bloom.
    \item A field of lavender flowers swaying in the breeze.
    \item A cozy coffee shop with a warm atmosphere and soft light.
    \item A stark, icy landscape with glaciers and frozen seas.
    \item A lush green valley surrounded by towering mountains.
\end{itemize}
\end{tcolorbox}
\vspace{-2mm}
\caption{Sample of 10 out of 100 captions used for generating the background with Stable Diffusion-XL~\cite{podell2023sdxlimprovinglatentdiffusion}}
    \label{tab:background-caption}
\end{minipage}
\end{table}

\end{document}